\documentclass[11pt]{article}

\usepackage[final]{acl}

\usepackage{times}
\usepackage{latexsym}

\usepackage[T1]{fontenc}
\usepackage[utf8]{inputenc}

\usepackage{microtype}

\usepackage{inconsolata}

\usepackage{graphicx}

\usepackage{array}
\usepackage{tabularx}
\usepackage{amsmath}

\title{Making Political Text Scaling Comparable: Infrastructure and
Hyperparameter Sensitivity for 17 Algorithms}

\author{
Patrick Parschan \\
Department of Media and Communication \\
LMU Munich \\
\texttt{p.parschan@lmu.de} \\
}

\begin{document}
\maketitle

\begin{abstract}
Computational text-based ideal point estimation (CT-IPE) methods are usually
compared as named algorithms, yet applying them involves numerous researcher
choices that configure how political text is turned into position estimates.
This paper argues that CT-IPE methods are better understood as configurable
measurement pipelines than as fixed estimators. Building on a large-scale
comparative experiment spanning 17 CT-IPE algorithms, 5,537 experimental
runs, and approximately 4.25 million left--right position estimates, I
describe the shared infrastructure that makes these heterogeneous methods
jointly executable and quantify how sensitive their estimates are to
alternative hyperparameter choices. Variance-partitioning and SHAP-based
sensitivity analyses show that, for most algorithms, hyperparameter
profiles explain little residual variance through a shared shift: 13 of
the 17 algorithms exhibit ICC values below .10. Where this profile-level
sensitivity is present, it is concentrated in a small number of
consequential researcher choices, most notably the selection of the
underlying language or embedding model, the seed keyword lists that
anchor the construct, and the number of topics.
\end{abstract}

\section{Introduction}
\label{sec:introduction}

Computational text-based ideal point estimation (CT-IPE), also known as
political text scaling, estimates numerical political positions from text,
most commonly on dimensions such as the left--right axis
\citep{carroll_ideal_2023,bafumi_practical_2005}. Over the past three decades,
the field has expanded from classical word-frequency models such as Wordscores \citep{laver_extracting_2003}
and Wordfish \citep{slapin_scaling_2008} to topic models \citep[e.g.,][]{vafa_text-based_2020}, embedding-based approaches \citep[e.g.,][]{rheault_word_2020}, and, more recently,
large language model (LLM)-based methods \citep[e.g.,][]{le_mens_positioning_2025}.
As this landscape has grown, methodological research has primarily compared algorithms
with respect to agreement with humans \citep{bruinsma_validating_2019, hjorth_computers_2015} or -- more limitedly -- with respect to agreement with other algorithms \citep{curini_scaling_2020}.
The named algorithm has consequently become the central unit of methodological investigation. 

However, applying a CT-IPE method involves considerably more than selecting an
algorithm. Researchers must also decide how political constructs are
operationalized, how texts are represented, how computational models are
configured, how external information is incorporated, and how political
positions are estimated. Although these choices are documented within
individual algorithms, they are rarely conceptualized as recurring
methodological decisions that span otherwise heterogeneous CT-IPE approaches.
As a result, conceptually similar researcher choices remain difficult to
identify and compare across algorithms. This raises two research questions:

\paragraph{RQ1:} How can heterogeneous CT-IPE algorithms be made jointly
executable and systematically comparable within a shared experimental
infrastructure?

\paragraph{RQ2:}~What patterns of researcher choice emerge from a systematic
comparison of heterogeneous CT-IPE algorithms?

Throughout, \emph{researcher choices} refers to the configurable decisions
involved in applying a CT-IPE method to data---hyperparameters, construct
anchors, model selection, and related pipeline settings---not to the author's
selection of which algorithms to include in the study.

To address these questions, this paper argues that CT-IPE methods are better
understood as configurable measurement pipelines than as fixed estimators.
Rather than comparing algorithms directly, I compare the researcher choices
that configure them. I build on a large-scale computational experiment that
executes 17 CT-IPE algorithms under a shared design, describe the software
infrastructure that makes these heterogeneous methods jointly runnable, and
quantify how strongly alternative hyperparameter settings influence the
resulting position estimates, using variance partitioning and SHAP-based
(SHapley Additive exPlanations, see \citet{lundberg_unified_2017})
sensitivity analysis.

The results show that, for most CT-IPE algorithms, alternative
hyperparameter profiles explain little of the remaining variance through a
shared shift within the evaluated specification space, and that where this
profile-level sensitivity is present, it is concentrated in a small number
of consequential researcher choices.

\section{Researcher Choices in CT-IPE}
\label{sec:researcher-choices}

CT-IPE methods are usually discussed
as named algorithms: Wordscores, Wordfish, Party Embeddings, TBIP, LaMP, or
other approaches.\footnote{For readability, I use only abbreviations (rather than full algorithm names) throughout the text. 
A complete list of all algorithms with their full names, corresponding abbreviations, and relevant papers is provided in Appendix~\ref{sec:app-algorithms}.} This algorithm-centered vocabulary is useful, but it hides
an important methodological feature of applied CT-IPE: algorithms are
configured through researcher choices. These choices determine how political
text is transformed into position estimates.

I use the term researcher choice or researcher decision to refer to a
theoretically or methodologically justifiable decision that may alter
estimated political positions. This definition is broader than the notion of a
software hyperparameter. A single researcher choice may be implemented through
multiple parameters within one algorithm and through different numbers or
combinations of parameters in other algorithms. Conversely, some
implementation parameters are purely technical and do not constitute
substantive researcher choices. This perspective is consistent with machine
learning (ML) and Natural Language Processing (NLP) research showing that
model behavior depends not only on the learning algorithm itself but also on
configurable choices throughout the modeling pipeline
\citep[e.g.,][]{bouthillier_accounting_2021,yang_hyperparameter_2020}. The
analyses below therefore treat hyperparameters as part of the measurement
procedure rather than as technical implementation details, and examine them
at the level of individual researcher choices.

\section{Methods}
\label{sec:methods}

\subsection{Comparative Experiment}
\label{sec:experiment}

To investigate researcher choices empirically, I build on a
large-scale comparative experiment that evaluates CT-IPE algorithms under a
shared design. The experiment comprises 17 unsupervised and semi-supervised
CT-IPE algorithms spanning four broad methodological types identified in the
systematic review of \citet{parschan_computational_2025}: word
frequency-based methods (Type~I), topic modeling-based methods (Type~II),
word embedding-based methods (Type~III), and LLM-based methods (Type~IV).
The included algorithms are the unsupervised and semi-supervised CT-IPE
methods covered by that review (see Appendix~\ref{sec:app-algorithms}).
Network-based approaches that infer ideology from non-textual signals
(e.g. \citet{barbera_tweeting_2015}) fall outside this text-based scope.

The algorithms are applied to estimate left--right positions in six corpora
consisting of party manifestos, parliamentary speeches, and political tweets
from Germany and the United States (see Table~\ref{tab:datasets}). These
textual genres are those for which CT-IPE methods have most frequently been
developed and applied \citep{parschan_computational_2025}.

\begin{table*}[t]
  \centering
  \small
  \begin{tabular}{llrrp{2.5cm}p{4.2cm}p{2.4cm}}
    \hline
    \textbf{Data genre} & \textbf{Country} & \textbf{N parties} &
    \textbf{N docs} & \textbf{Timespan} & \textbf{Political context} &
    \textbf{Reference data source} \\
    \hline
    Party manifestos & US & 2 & 12 &
    2004-11-01 until 2024-12-31 &
    Six US presidential elections from 2004 to 2024 &
    \citet{lehmann_manifesto_2025} \\
    Legislative speeches & US & 2 & 910 &
    2020-09-01 until 2020-10-31 &
    Two months before the US presidential election 2020 &
    \citet{judd_congressional-record_2019} \\
    Political tweets & US & 2 & 2,445 &
    2024-10-01 until 2024-11-05 &
    Final five weeks before the 2024 U.S. presidential election &
    \citet{balasubramanian_public_2024} \\
    Party manifestos & Germany & 6 & 12 &
    2021-09-01 until 2025-02-28 &
    Two federal elections in Germany in 2021 and 2025 &
    \citet{lehmann_manifesto_2025} \\
    Legislative speeches & Germany & 6 & 713 &
    2021-05-01 until 2021-06-30 &
    Two months before the summer break before the German federal election
    2021 &
    \citet{lange_speakger_2023} \\
    Political tweets & Germany & 6 & 3,077 &
    2017-08-01 until 2017-09-30 &
    Two months before the German federal election 2017 &
    \citet{kratzke_btw17_2017} \\
    \hline
  \end{tabular}
  \caption{Datasets overview. Number of parties and documents per corpus.}
  \label{tab:datasets}
\end{table*}

Section~\ref{sec:implementation} describes how the shared implementation
pipeline makes these heterogeneous methods jointly executable under one
comparative design.

The resulting experiment comprises 5,537 attempted runs, of which 4,565
completed successfully, producing approximately 4.25 million political
position estimates. These provide the empirical basis for the analyses
presented below. Summary statistics for the runs can be found in
Appendix~\ref{sec:app-runs}. Run counts differ considerably across
algorithms because hyperparameter grids differ in size and because runtime
costs limited the number of evaluated configurations for computationally
expensive methods. Each run corresponds to a configuration of
implementation parameter values drawn from the experimental grid, which
together instantiate a specific configuration of researcher choices.

The evaluated specification space was designed to be as comprehensive as
possible: implementation parameters and their candidate values were
systematically derived from the CT-IPE literature review of
\citet{parschan_computational_2025}. Different experimental designs or
additional researcher choices may nevertheless yield different
sensitivity profiles.

\subsection{Code Implementation}
\label{sec:implementation}

Making 17 heterogeneous CT-IPE methods executable under a shared comparative
design required substantial software engineering. The methods were originally
developed as standalone tools with corpus-specific assumptions, heterogeneous
APIs, and incompatible output formats. The project therefore executes all
compared algorithms through algorithm-specific Python wrappers that connect
their existing implementations to the shared corpora, metadata, and
hyperparameter grid. The wrappers are not intended to reimplement the
algorithms or standardize their internal estimation logic; their purpose is to
make heterogeneous methods jointly runnable while preserving native
implementations.

Native algorithm outputs are retained as produced during inference. Only
during analysis are document-level estimates aggregated to party--source cells
where necessary and estimates robustly standardized to a common scale.
The sensitivity analysis below uses these analysis-time transforms but
leaves each run's native polarity unchanged.\footnote{Polarity is the
orientation of a left--right scale: which end corresponds to left and
which to right. Several unsupervised estimators identify the axis only
up to sign, so otherwise similar runs can place left parties at opposite
ends. Native polarity is this sign as the algorithm emitted it.} A robustness check
additionally aligns polarity per run using fixed anchor parties, so that
the right-anchor party receives the larger value
(Section~\ref{sec:overall-sensitivity}). This keeps the original method
outputs distinct from the quantities that enter the variance
decomposition.

A fully literal replication of every original implementation was not always
possible: some methods were developed for specific corpora, model versions,
APIs, preprocessing choices, or hardware environments. Where exact replication
would have made the broader comparison infeasible, I introduced bounded
modifications of the original setup.\footnote{These modifications are documented in \texttt{MODIFICATIONS.md} in the online repository (Appendix~\ref{sec:app-online-supplementary}).}

Three engineering lessons proved especially consequential for comparative
CT-IPE work. First, a thin wrapper around each native estimator is more
maintainable than a forced common API: methods differ too sharply in inputs
(document--term matrices, embeddings, prompts, pairwise rankings) for a
single interface, yet they can still share corpora, metadata, logging, and
result storage. Second, configuration validation must be treated as part of
the experimental record. Unknown keys in the grid used to be dropped
before execution instead of failing; configured and executed conditions
could therefore silently diverge. Unknown keys are now rejected at
planning time so that the written grid and the runs stay aligned
(see the Limitations section). Third, resource heterogeneity is itself a
researcher constraint: runtimes range from seconds to tens of minutes per
configuration, so equal-sized grids are often infeasible and uneven executed
samples are an expected property of a joint design rather than a defect of
it (Appendix~\ref{sec:app-runs}).

The CPU system used for running and developing the code was equipped with
32~vCPUs (AMD EPYC 7H12) and 125\,GB RAM. The GPU system used for inference
was equipped with an RTX PRO 6000 with 96\,GB VRAM and the same CPU and RAM
settings. The machines were rented via \url{https://runpod.io}. Costs for
development and inference amounted to \$1,125.

Taken together, this shared execution layer answers RQ1: heterogeneous
CT-IPE algorithms become jointly executable and systematically comparable
through algorithm-specific wrappers around preserved native
implementations, a shared configuration grid, and analysis-time
standardization that is kept separate from native outputs.

\section{Results}
\label{sec:results}

Section~\ref{sec:results} addresses RQ2. The empirical analysis proceeds in
two steps: a variance decomposition that quantifies profile-level
(common-direction) sensitivity, and a SHAP-based attribution of that
shared shift to individual researcher choices.

\subsection{Estimand and Estimation}
\label{sec:estimand}

Estimates are analyzed in a defined experimental context (see Table~\ref{tab:datasets}). Context
comprises country, text genre (data source), party,
result granularity,\footnote{Whether the algorithm estimates positions
at the document level or directly at the actor (party) level.} and
time-period;\footnote{Some semi-supervised algorithms take time-slicing as an input.} where an algorithm was run with multiple
random number seeds, the seed is likewise treated as a contextual
factor. Document- and actor-level scores are aggregated to the party in each
context and, within each algorithm, robustly standardized using the
median and interquartile range. The unit of analysis is then one
standardized left--right score for a party in a given context under one
hyperparameter configuration.

The analysis is conducted separately for each algorithm. The quantity of
interest is how strongly alternative researcher choices move party scores
once the experimental setting is held constant. I group those choices into
cleaned hyperparameter profiles: the substantive decisions that define a
configuration, after setting aside random seeds that capture only
estimation noise and parameters that merely restate the corpus context
(for example, country-specific prompt prefixes). Let $y_{ipc}$ denote the
standardized left--right score for party $p$ under profile $i$ in
context $c$. I estimate
\begin{equation}
y_{ipc} = \mathbf{x}_{c}^{\top}\boldsymbol{\beta} + u_{i} + \varepsilon_{ipc},
\end{equation}
where $\mathbf{x}_{c}$ accounts for the context factors that vary for that
algorithm, $u_{i}$ is a shared shift associated with profile $i$ (variance
$\tau^{2}$), and $\varepsilon_{ipc}$ is leftover variation within profiles
(variance $\sigma^{2}$). Hyperparameter sensitivity is the share of this
remaining variation that lies between profiles,
\begin{equation}
\rho = \frac{\tau^{2}}{\tau^{2}+\sigma^{2}},
\end{equation}
the intraclass correlation (ICC). Values near zero mean that, in the same
setting, differences between hyperparameter profiles account for little of
the remaining variance through a shared shift; values near one mean that
those differences account for much of the remaining variance after
contextual adjustment. Because $u_{i}$ is a common shift, the ICC captures
researcher-choice effects that move parties in the same direction.
Effects that reorder parties or reverse the scale's polarity are absorbed
into the residual; they are examined separately in the polarity-alignment
robustness check (Section~\ref{sec:overall-sensitivity}) and the party
landscapes (Section~\ref{sec:detailed-analysis}). A low ICC therefore does
not imply that researcher choices cannot rearrange relative party
positions.

The ICC describes the
executed specification space. It does not identify the causal effect of
any single parameter, and algorithms with few configurations yield noisier
estimates. The ICC therefore asks how much of the remaining
variation, after the experimental setting is held constant, is associated
with hyperparameter profiles as a whole. To unpack that bundle into
individual researcher choices, I fit an ANOVA-style linear model. The
dependent variable is that same context-adjusted party score; the
independent variables are the individual hyperparameters that define the
cleaned profiles, not every implementation knob in the grid. That linear
model uses hyperparameter main effects rather than
party-by-hyperparameter interactions, so the attribution likewise concerns
common-direction movement rather than party-specific reordering. SHAP values
then show how much each of those parameters contributes.
Uncertainty is reported as 95\% intervals from 1,000 bootstrap resamples
of hyperparameter profiles.

\subsection{Profile-Level Sensitivity}
\label{sec:overall-sensitivity}

\begin{figure*}[t]
  \centering
  \includegraphics[width=\textwidth]{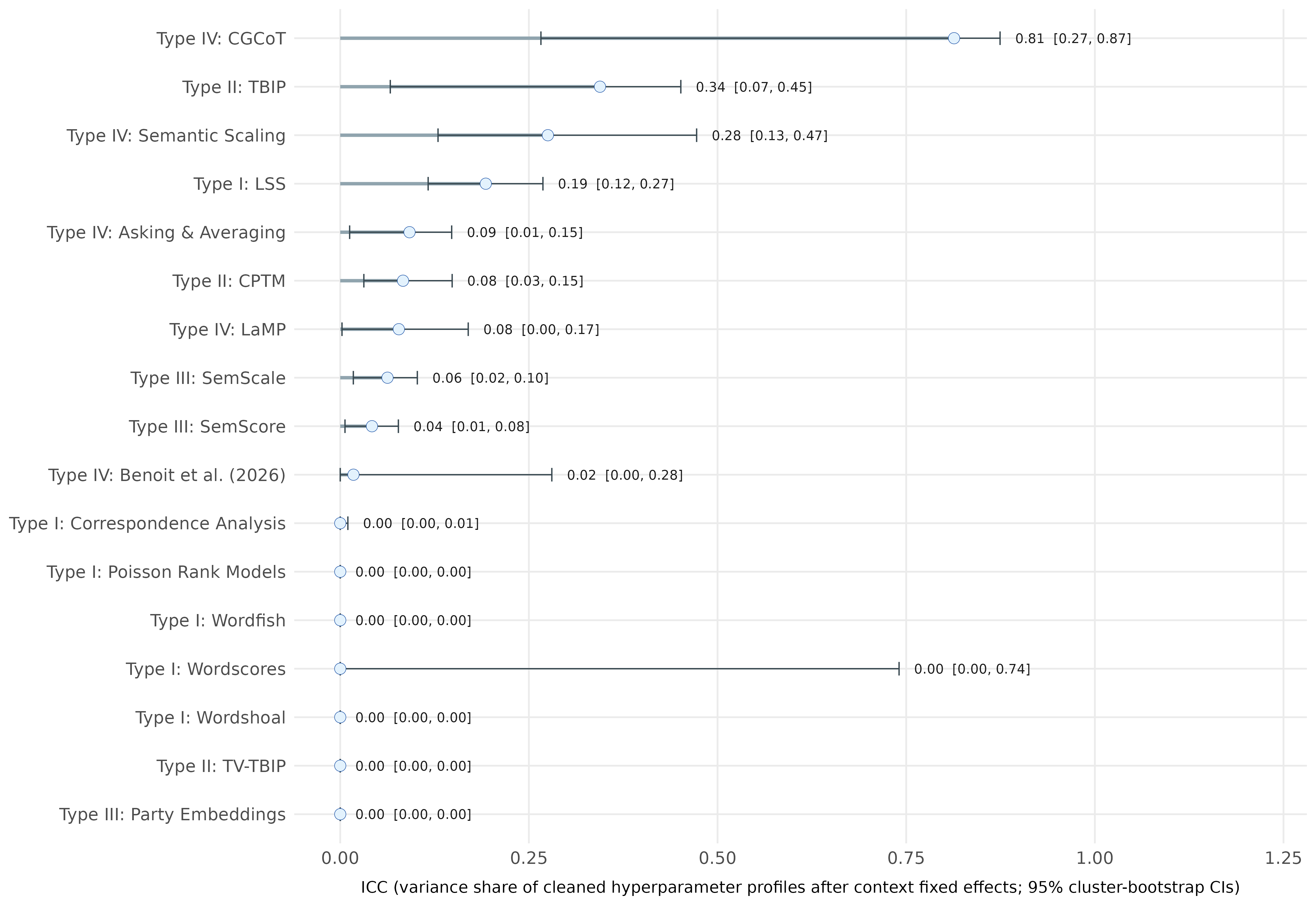}
  \caption{Profile-level (common-direction) hyperparameter sensitivity
    across the 17 CT-IPE algorithms. Points show the ICC: the proportion of
    variation in context-residualized political position estimates
    attributable to a shared shift associated with alternative cleaned
    hyperparameter profiles; horizontal lines and bracketed values give
    95\% percentile intervals from a cluster bootstrap over cleaned
    profiles (1,000 replicates). Party-specific reordering is not captured
    by this quantity.}
  \label{fig:icc}
\end{figure*}

Figure~\ref{fig:icc} summarizes profile-level (common-direction)
sensitivity: the ICC share of remaining variance associated with a shared
shift under each cleaned hyperparameter profile. It is not a measure of
every way researcher choices can change estimates, because party-specific
reordering is absorbed into the residual.

The results reveal that this shared-shift variance is small for most
algorithms. Thirteen of the 17 methods exhibit ICC values below .10; for
seven of them, the estimated share is essentially zero. For these
algorithms, alternative researcher choices within the evaluated design
explain little residual variance through a common intercept. Four algorithms
show non-negligible profile-level sensitivity: LSS (.19), Semantic Scaling
(.28), and TBIP (.34) occupy an intermediate range, while CGCoT (.81) is
the only algorithm for which differences between hyperparameter settings
dominate the remaining variation after contextual adjustment. As discussed
in Section~\ref{sec:detailed-analysis}, the CGCoT estimate rests on very
few executed configurations and should be interpreted with particular
caution.

Low ICC values are not an artifact of empty or trivial grids. Several
methods with near-zero ICCs still vary many cleaned profiles and
substantive parameters: Wordscores (ICC~$=.00$) has 136 cleaned profiles
across seven retained parameters; Wordfish ($.00$) has 72 profiles and six
parameters; Correspondence Analysis ($.00$) has 64 profiles; and Party
Embeddings ($.00$) has 113 profiles across six parameters
(Appendix~\ref{sec:app-runs}). By contrast, CGCoT's high ICC rests on only
four cleaned profiles. A near-zero ICC in this design therefore means that
profile differences explain little of the context-adjusted variance through
a shared shift despite nontrivial researcher-choice variation, not that
little was varied.

The bootstrap intervals in Figure~\ref{fig:icc} support this reading while
making its uncertainty explicit. The intervals of the four algorithms with
non-negligible ICCs exclude zero, but they are wide where the executed
grids are small: CGCoT's interval spans [.27, .87] with only four cleaned
profiles, and the interval of \citet{benoit_using_2026} ([.00, .28], six
profiles) is compatible with both negligible and moderate profile-level
sensitivity. For most algorithms with near-zero point estimates, the
intervals are tight around zero.
Wordscores is the exception ([.00, .74]): its point estimate is zero, but
a small share of configurations produces extremely unstable estimates
(some fully degenerate reference-rescaling runs were already excluded during
data preparation), so resamples that overweight these configurations
yield large variance shares. Its low point estimate should therefore be read
as ``hyperparameters explain little of a noisy signal''; its profile-level
sensitivity cannot be characterized precisely because of these unstable
configurations.

A second robustness concern is polarity. For several algorithms, left and
right can swap from run to run. The ICC only counts a shared shift that
moves parties in the same direction, so those flips are not registered as
profile-level sensitivity. They instead add leftover scatter, which can
shrink the ICC and hide a genuine common-direction effect. Re-estimating all ICCs
after applying an anchor-based polarity alignment,
which sign-flips every run whose fixed anchor parties appear in reversed
order, leaves the substantive picture intact
(Appendix~\ref{sec:app-polarity}): the near-zero algorithms remain at
essentially zero although between 19\% and 53\% of their runs are
mirrored, so their low sensitivity is not an artifact of sign
indeterminacy. Two shifts are informative. SemScale rises from .06 to .16,
indicating that polarity flips of its embedding-based scale masked
genuine profile-driven variance; its headline value understates its
sensitivity. Conversely, CGCoT falls from .81 to .36 and LSS from .19 to
.10: their polarity is substantively defined by prompts and seed keywords,
so part of their measured sensitivity consists of configurations that
reverse the direction of the scale outright rather than shifting positions
along it.\footnote{Seed-keyword lists are in \texttt{configs/experiment\_grid.yaml} on the online repository; LLM prompt files are in each algorithm's \texttt{prompt\_configs} folder.}

However, Figure~\ref{fig:icc} should not be interpreted as a ranking of
methodological quality. High profile-level sensitivity does not imply that an
algorithm is inferior, nor does a near-zero ICC imply greater validity.
Instead, the results quantify the extent to which alternative, theoretically
or methodologically justifiable researcher choices produce a shared shift
in political position estimates within the evaluated comparison framework.
Sensitivity to construct-defining hyperparameters, such as seed keywords,
can even be desirable: those choices are how researchers steer which
variation in the text is treated as the relevant signal for the construct.

A first answer to RQ2 is therefore that, for most of the sampled named
algorithms, hyperparameter profiles explain little residual variance
through a shared shift, and that only a small subset exhibits substantial
profile-level sensitivity. The following section examines which
individual researcher choices these differences arise from.

\subsection{Which Researcher Choices Drive Sensitivity?}
\label{sec:detailed-analysis}

\begin{figure*}[t]
  \centering
  \includegraphics[width=\textwidth]{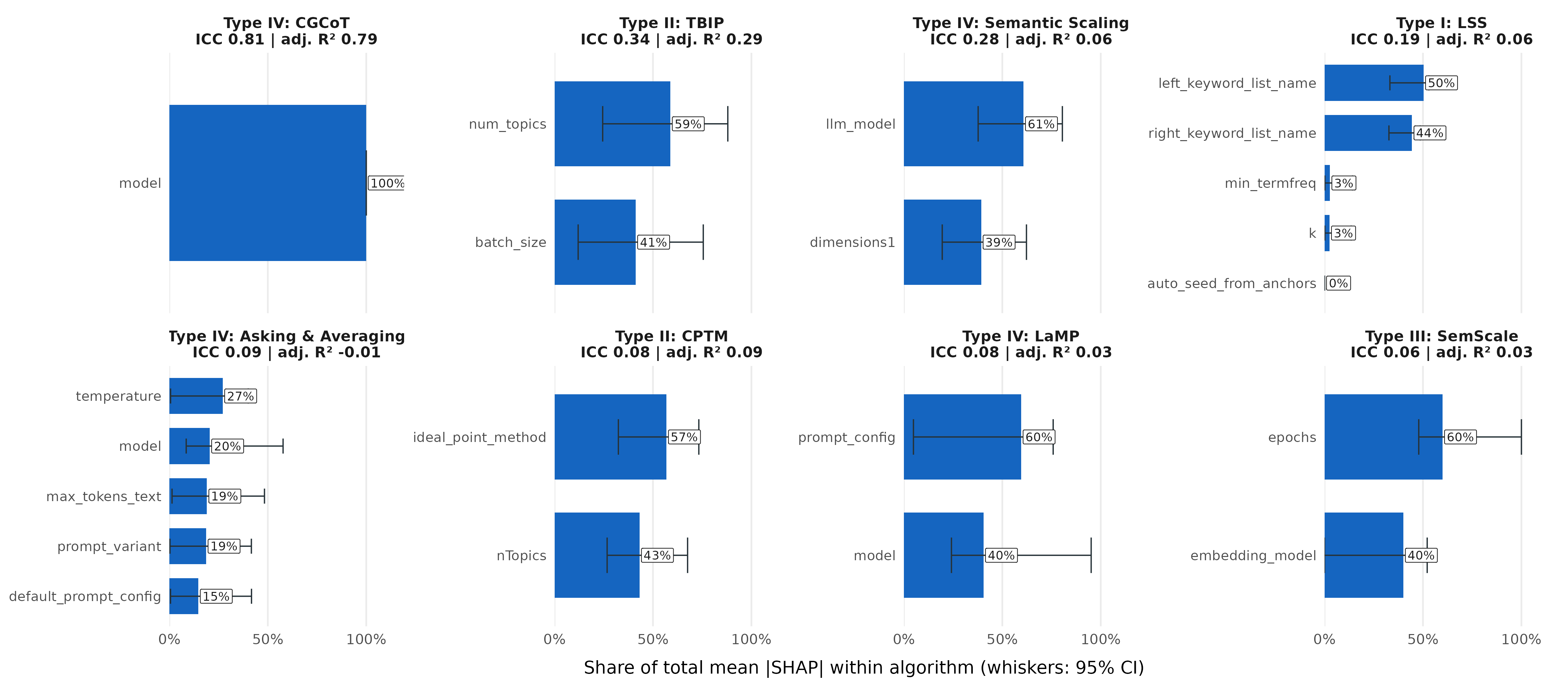}
  \caption{Within-algorithm distribution of profile-level hyperparameter
    sensitivity for the eight algorithms with an ICC of at least .05. Bars
    show the share of total mean absolute SHAP value attributable to each
    varied implementation parameter; whiskers give 95\% percentile
    intervals from the cluster bootstrap over cleaned profiles (1,000
    replicates); facets are ordered by the ICC in
    Figure~\ref{fig:icc}. Facet headers also
    report the adjusted $R^{2}$ of that linear model; where it is near
    zero, shares are exploratory rankings rather than precise partitions of
    shared-shift variance.}
  \label{fig:decisions}
\end{figure*}

The profile-level ICC does not identify which choices produce the shared
shift. To answer that question, I use the ANOVA-style linear model
described above and SHAP values to attribute leftover variation to the
implementation parameters that varied within the executed specification
space. Because that model uses hyperparameter main effects rather than
party-by-hyperparameter interactions, the shares likewise attribute
common-direction sensitivity. The resulting importances are normalized within each
algorithm and provide an exploratory attribution of hyperparameter-driven
variation to individual parameters.
Figure~\ref{fig:decisions} shows these within-algorithm importance shares
for the eight algorithms with a profile-level ICC of at least .05;
Appendix~\ref{sec:app-shap} reports the corresponding profiles for all 17
algorithms. Two diagnostics bound what the shares can support, and both are
reported in the facet headers. First, the shares are proportions of each
algorithm's own -- possibly negligible -- shared-shift variance, so they
must be read together with the ICC. Second, the adjusted $R^{2}$ of
this linear model indicates how much of the context-adjusted
variation the attribution model itself captures: it is substantial only for
CGCoT (.79) and TBIP (.29), whereas for LSS and Semantic Scaling
($\approx.06$) and the remaining algorithms the shares rank candidate
parameters within a weakly explained outcome rather than partitioning it
precisely.

A first observation is that, where profile-level sensitivity exists, it is
highly concentrated: for every algorithm in Figure~\ref{fig:decisions}, one
or two parameters account for the bulk of the attributed importance. For
CGCoT and the approach of \citet{benoit_using_2026}, the choice of the
underlying LLM is the only consequential parameter; it was also the only
parameter varied in the experiment (see Appendix~\ref{sec:app-runs}).
Given the prompt insensitivity reported by \citet{benoit_using_2026} and
the resource demand of these two algorithms, I opted not to make prompt
variation a major focus of the experiment for CGCoT and
\citet{benoit_using_2026}.
Prompt variation was included for LaMP and Asking \& Averaging, which
expose the prompt as a hyperparameter.
Once prompt wording is set aside, LLM model choice is the remaining
parameter that the attribution flags as consequential for those
algorithms.\footnote{Apart from decoding knobs such as temperature and
top-$p$ sampling.}
For Semantic Scaling,
the underlying NLI model and the natural-language definition of the scaling
dimension (the entailment hypotheses) dominate, although the bootstrap
interval for the latter is wide ([.20, .62]). For TBIP, sensitivity derives from the number of topics and the
training batch size, and for LSS from the seed keyword lists that define
the ideological poles; for LSS, the dominance of the seed keywords is
corroborated independently by the landscape analysis below. Across the
evaluated specification space, the choices flagged as consequential thus
reduce to a short list: which model is used, which seed keywords anchor the
construct, and how many topics are estimated -- read as a robust ranking
where that linear model fits weakly rather than as an exact variance
decomposition. Conversely, the preprocessing parameters often emphasized for
classical frequency-based methods, such as term- and document-frequency
thresholds, show no measurable influence on context-residualized estimates
within the evaluated grids: Wordscores, Wordfish, Correspondence Analysis,
and Wordshoal all exhibit ICC values of zero.

\begin{figure*}[t]
  \centering
  \includegraphics[width=0.88\textwidth]{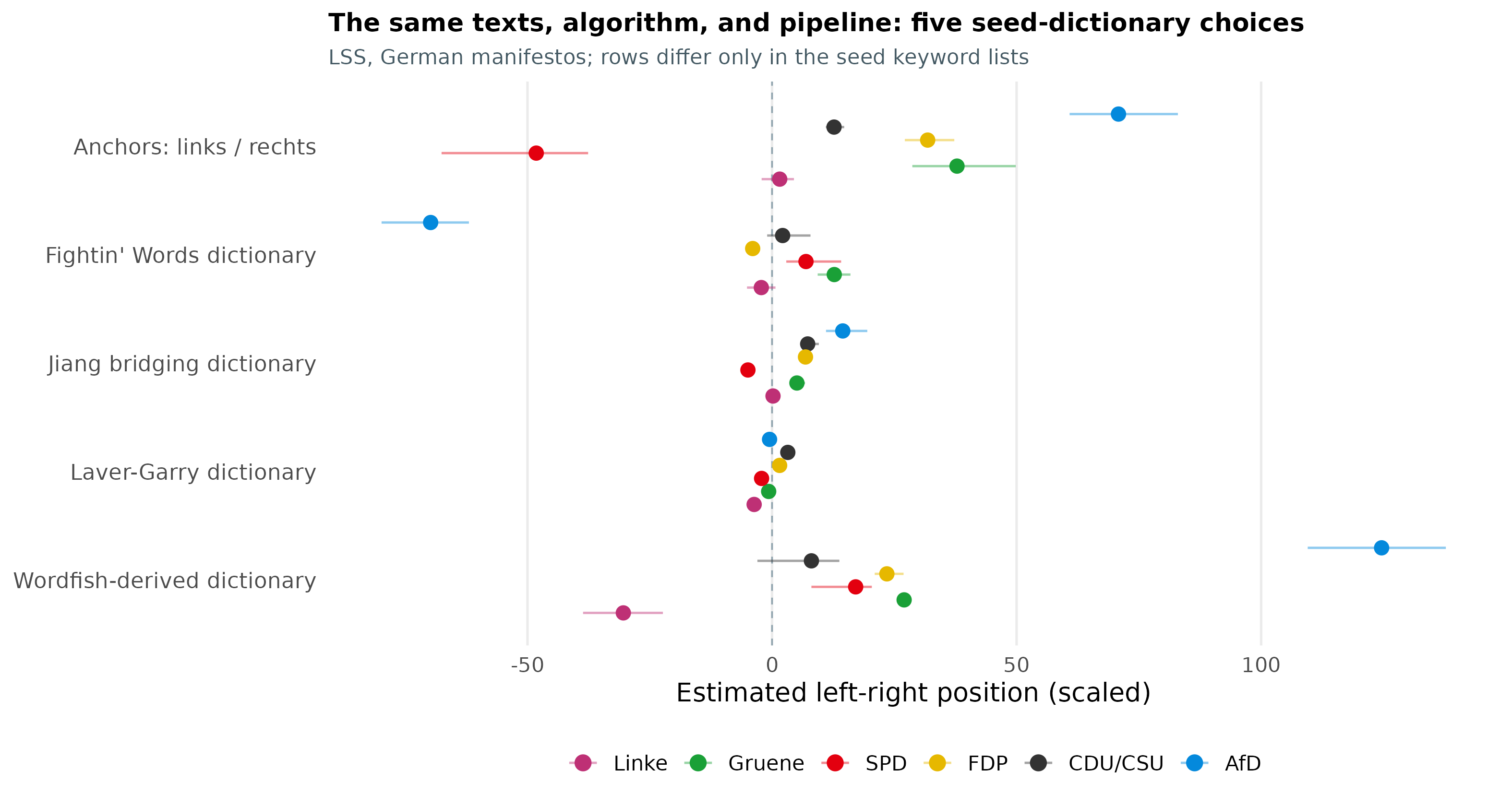}
  \caption{Party landscapes under alternative LSS seed-keyword choices
    (German manifestos). Points are mean scaled left--right estimates per
    party within each keyword family; whiskers span the range of
    specification means within that family. Rows differ only in the seed
    keyword lists (and the auto-seeding mode coupled to them in the
    executed grid). A pooled row of additional generic anchor pairs is
    omitted for readability.}
  \label{fig:landscape}
\end{figure*}

Figure~\ref{fig:landscape} makes the practical consequence of such
concentrated sensitivity concrete. Holding the texts, algorithm, and
pipeline fixed, it maps the six German parties under five alternative LSS
seed-keyword choices. With the bare German anchors \textit{links}/\textit{rechts},
the AfD is placed far right and the SPD far left; with the Fightin' Words \citep{monroe_fightin_2008}
dictionary the AfD flips to the far left; with the Wordfish-derived \citep{slapin_scaling_2008}
dictionary it is again extreme right, while the Laver--Garry \citep[see][]{laver_extracting_2003} and \citet{jiang_bridging_2024}
dictionaries compress nearly all parties toward the center. The
figure is an illustration for one algorithm and corpus, not a comparative
ranking of dictionaries; in the executed LSS grid the keyword family also
determines whether auto-seeding from anchors is used, so the rows show the
joint effect of those coupled choices.
Appendix~\ref{sec:app-landscape} reports the same presentation for
Semantic Scaling and TBIP, with Wordfish as a low-sensitivity contrast.

LLM-based approaches require somewhat different interpretation. Semantic
Scaling relies on natural language inference (NLI) rather than prompting
generatively trained LLMs, so construct specification is operationally
distinct from prompt engineering
\citep{laurer_less_2023,halterman_codebook_2026}. Within the
evaluated specification space, Semantic Scaling exhibits greater
profile-level sensitivity (.28) than the generative approaches LaMP (.08),
Asking \& Averaging (.09), and \citet{benoit_using_2026} (.02), driven by
the underlying NLI model and the natural-language definition of the scaling
dimension. For the generative LLM methods, even where prompt-related
parameters were included in the experiment and accounted for a
recognizable share of the within-algorithm profile (LaMP and A\&A), they
partition a profile-level sensitivity that is small in absolute terms. This
accords with the prototyping analysis in the supplementary material of
\citet{benoit_using_2026}, which finds that systematic small changes in
scoring-prompt wording rarely alter position scores and are ``not critical
in this application.''

The primary exception within this group is CGCoT, whose comparatively high
sensitivity should be interpreted cautiously. Owing to its substantially
higher computational cost, fewer configurations could be
evaluated than for most other algorithms
(Appendix~\ref{sec:app-runs}). Moreover, CGCoT was developed for short
political texts; manifesto corpora were therefore excluded from its
executed grid, and the algorithm was evaluated on speeches and tweets
only. Its pairwise
ranking procedure also requires more than two political actors, rendering it
inapplicable to the two-party U.S. datasets. Consequently, only 8 of the 16
attempted specifications produced valid results, and the estimated
sensitivity rests on this very small executed grid.

Similar caution applies to other algorithms with small executed grids or
low run counts (Appendix~\ref{sec:app-runs}); comparisons involving them
should be read descriptively rather than causally. SemScale (.06) and
SemScore (.04) are not in that thin-grid group: their low ICCs rest on
substantial executed samples, including in-domain embedding training
(Appendix~\ref{sec:app-runs}, see also the Limitations section).

\enlargethispage{4\baselineskip}
\section{Conclusion}
\label{sec:conclusion}

This paper argued that CT-IPE methods are better understood as
configurable measurement pipelines than as fixed algorithms, and compared
the researcher choices that configure 17 methods in a shared experiment.
Within the evaluated specification space, hyperparameter profiles account
for less than 10\% of the context-adjusted variance through a shared shift
for 13 of the 17 algorithms; the broader pattern remains after
anchor-based polarity alignment. Where this profile-level sensitivity is
present, it concentrates in a few decisions---the underlying language or
embedding model, the seed keywords that anchor the construct, and the
number of topics---rather than spreading across the pipeline.

The named algorithm therefore remains a reasonable unit of comparison
where those shared shifts are small. A few construct-defining choices can
still steer the estimates, and those choices deserve explicit reporting.
Party-specific reordering lies outside the ICC estimand.

\section*{Limitations}
\label{sec:limitations}

Several limitations should be acknowledged. The reported analysis
characterizes the researcher choices varied in the present comparative
experiment rather than every conceivable implementation choice, and results
may change as additional algorithms, corpora, or methodological decisions
are incorporated.

The ICC reported here is a profile-level shared-shift quantity.
Party-specific reordering is absorbed into the residual and is illustrated
only for selected algorithms and corpora, so a near-zero ICC should not be
read as immunity of the party landscape to researcher choices.

The experimental grids are also uneven across algorithms. That imbalance is
largely unavoidable: the methods expose different numbers of
hyperparameters, and resource constraints limited how many configurations
could be run for the more expensive ones. Cross-algorithm contrasts of
sensitivity should therefore be read descriptively, especially where
executed samples are thin.

In addition, a post-hoc audit of the execution pipeline found that
configuration validation originally dropped unknown YAML keys instead of
failing. A written grid and the executed runs could therefore silently
diverge. Unknown keys are now rejected at planning time. The sensitivity
estimates reported here are for the executed specification space. Where
pretrained embeddings are used, the accompanying training knobs
(window size, epochs, and the like) are placeholders and do not constitute
additional trained variation.

Future work should therefore broaden the evaluated specification space,
investigate additional forms of researcher choice, and examine how
hyperparameter sensitivity relates to substantive conclusions drawn from
political text scaling.

\bibliography{custom}

\clearpage
\onecolumn
\raggedbottom
\appendix

\section{Included Algorithms and Relevant Papers}
\label{sec:app-algorithms}

\noindent
\begin{minipage}{\textwidth}
  \centering
  \footnotesize
  \begin{tabular}{p{3.4cm}p{3.2cm}p{4.6cm}p{1.6cm}p{0.9cm}}
    \hline
    \textbf{Relevant references} & \textbf{Algorithm name} &
    \textbf{Comment} & \textbf{Supervision} & \textbf{Type} \\
    \hline
    \citep{laver_extracting_2003,lowe_understanding_2008,martin_robust_2008,beauchamp_using_2012} &
    Wordscores (WS) &
    The comparison includes derivations and rescaling ideas made after the
    native paper. &
    Semi-supervised & Type I \\
    \citep{slapin_scaling_2008,riesch_wordkrill_2025} &
    Wordfish (WF) &
    Wordkrill by \citet{riesch_wordkrill_2025} is a recent extension of
    Wordfish into more than one dimension. &
    Unsupervised & Type I \\
    \citep{lauderdale_measuring_2016} &
    Wordshoal (WSH) &
    Two-stage model: debate-level Wordfish scaling aggregated into
    speaker positions. &
    Unsupervised & Type I \\
    \citep{gabel_putting_2000,lowe_scaling_2016,saltzer_finding_2022} &
    Correspondence Analysis (CA) &
    \citet{saltzer_finding_2022} is an application paper. &
    Unsupervised & Type I \\
    \citep{jentsch_poisson_2021,jentsch_time-dependent_2020} &
    Poisson Reduced-Rank Models (PRM) &
    \citet{jentsch_time-dependent_2020} is a version of PRM that includes
    time as a variable. The comparison only includes
    \citet{jentsch_poisson_2021}, because the repository for PRM only covers
    that. &
    Unsupervised & Type I \\
    \citep{watanabe_latent_2021,watanabe_left-right_2015,warode_mapping_2025} &
    Latent Semantic Scaling (LSS) &
    \citet{watanabe_left-right_2015} is a blog post by the LSS author going
    into how LSS can be used for left-right scaling.
    \citet{warode_mapping_2025} is an applied paper. &
    Semi-supervised & Type I \\
    \citep{vafa_text-based_2020} &
    Text-Based Ideal Points (TBIP) & &
    Unsupervised & Type II \\
    \citep{hofmarcher_revisiting_2025,vafa_text-based_2020,gentzkow_measuring_2019} &
    Time-Varying Text-Based Ideal Points (TV-TBIP) &
    \citet{hofmarcher_revisiting_2025} combine ideas from
    \citet{vafa_text-based_2020} and \citet{gentzkow_measuring_2019} to
    include time into the TBIP algorithm. &
    Unsupervised & Type II \\
    \citep{van_der_zwaan_validating_2016,fang_mining_2012} &
    Cross-Perspective Topic Models (CPTM) &
    \citet{van_der_zwaan_validating_2016} build upon ideas from
    \citet{fang_mining_2012} to improve CPTM for the IPE task. &
    Unsupervised & Type II \\
    \citep{rheault_word_2020} &
    Party Embeddings (PE) & -- &
    Unsupervised & Type III \\
    \citep{nanni_political_2022,glavas_unsupervised_2017} &
    SemScale &
    The unsupervised version of SemScore. &
    Unsupervised & Type III \\
    \citep{nanni_political_2022,glavas_unsupervised_2017} &
    SemScore &
    The semi-supervised version of SemScale. &
    Semi-supervised & Type III \\
    \citep{wu_large_2023} &
    Language Model Pairwise Comparison (LaMP) & -- &
    Semi-supervised & Type IV \\
    \citep{wu_concept-guided_2024} &
    Concept-Guided Chain-of-Thought (CGCoT) &
    Building on the LaMP paper by \citet{wu_large_2023}. &
    Semi-supervised & Type IV \\
    \citep{burnham_semantic_2024} &
    Semantic Scaling (SS) & -- &
    Semi-supervised & Type IV \\
    \citep{le_mens_positioning_2025} &
    Asking \& Averaging (AA) & -- &
    Semi-supervised & Type IV \\
    \citep{benoit_using_2026} &
    Benoit '26 &
    LLM-based scaling through natural language understanding; the approach
    carries no established name and is referred to by its authors here. &
    Semi-supervised & Type IV \\
    \hline
  \end{tabular}
  \captionof{table}{Overview of all reviewed CT-IPE algorithms. The main
    paper relevant for the algorithm and the comparison implemented in this
    study is always mentioned first.}
  \label{tab:algorithms}
\end{minipage}

\section{Overview of Algorithm Runs}
\label{sec:app-runs}

\noindent
\begin{minipage}{\textwidth}
  \centering
  \small
  \begin{tabular}{lrrrrrr}
    \hline
    \textbf{Algorithm} & \textbf{Runs} & \textbf{Successful} &
    \textbf{Runtime} & \textbf{RAM (GB)} & \textbf{Profiles} & \textbf{Params} \\
    \hline
    Wordscores & 672   & 616 & 24s  & 1.2  & 136 & 7 \\
    Wordfish   & 480   & 304 & 22s  & 11.6 & 72  & 6 \\
    CA         & 480   & 480 & 15s  & 2.6  & 64  & 5 \\
    Wordshoal  & 336   & 184 & 27s  & 7.3  & 20  & 5 \\
    PRM        & 90    & 90  & 163s & 4.1  & 10  & 4 \\
    LSS        & 1,080 & 832 & 16s  & 4.8  & 312 & 9 \\
    TBIP       & 48    & 48  & 558s & 4.2  & 8   & 3 \\
    TV-TBIP    & 48    & 44  & 303s & 2.5  & 20  & 5 \\
    CPTM       & 192   & 192 & 83s  & 11.0 & 32  & 3 \\
    PE         & 668   & 666 & 10s  & 5.8  & 113 & 6 \\
    SemScale   & 192   & 191 & 120s & 7.6  & 32  & 6 \\
    SemScore   & 440   & 440 & 128s & 7.6  & 150 & 7 \\
    LaMP       & 378   & 126 & 580s & 28.1 & 14  & 2 \\
    CGCoT      & 16    & 8   & 35m  & 6.6  & 4   & 1 \\
    SS         & 180   & 115 & 38s  & 55.3 & 60  & 3 \\
    A\&A       & 192   & 192 & 250s & 3.2  & 24  & 7 \\
    Benoit '26 & 45    & 37  & 33m  & 6.4  & 6   & 1 \\
    \hline
  \end{tabular}
  \captionof{table}{Per-algorithm overview: number of attempted runs,
    successful runs, mean runtime, peak RAM, and the number of cleaned
    hyperparameter profiles and retained substantive parameters entering
    the sensitivity analysis (Section~\ref{sec:estimand}). Runs fail or are
    skipped when an algorithm is not applicable to a corpus or a
    configuration does not complete. Total position estimates across all
    algorithms: 4,252,357. Total attempted runs: 5,537; successful runs:
    4,565.}
  \label{tab:runs}
\end{minipage}

\section{Polarity-Alignment Robustness Check}
\label{sec:app-polarity}

Table~\ref{tab:polarity} reports the cleaned-profile ICC of
Figure~\ref{fig:icc} after per-run anchor-based polarity alignment:
every run whose fixed anchor parties (Gr\"une vs.\ AfD for
Germany, Democrat vs.\ Republican for the US) appear in reversed order is
sign-flipped before the per-algorithm robust standardization and the
variance decomposition are recomputed. Runs for which no anchor order can
be established (missing anchor party or tie) are excluded. The flipped share is a descriptive
polarity-instability diagnostic; it mixes arbitrary sign indeterminacy
(unsupervised scales) with substantive choice-induced reversal (where
prompts, seed keywords, or entailment hypotheses define the scale's
direction). Alignment therefore removes a nuisance for the former group
but part of the measured researcher-choice effect for the latter.

\noindent
\begin{minipage}{\textwidth}
  \centering
  \small
  \begin{tabular}{lrrrr}
    \hline
    \textbf{Algorithm} & \textbf{ICC} & \textbf{ICC aligned} &
    \textbf{Flipped (\%)} & \textbf{Unalignable (\%)} \\
    \hline
    CGCoT      & .81 & .36 & 12.5 & 0.0  \\
    TBIP       & .34 & .30 & 54.2 & 0.0  \\
    SS         & .28 & .38 & 48.7 & 32.2 \\
    LSS        & .19 & .10 & 50.5 & 0.0  \\
    A\&A       & .09 & .09 & 6.8  & 0.0  \\
    CPTM       & .08 & .02 & 48.0 & 22.9 \\
    LaMP       & .08 & .08 & 0.0  & 0.0  \\
    SemScale   & .06 & .16 & 40.3 & 0.0  \\
    SemScore   & .04 & .04 & 0.0  & 0.0  \\
    Benoit '26 & .02 & .02 & 0.0  & 2.7  \\
    CA         & .00 & .00 & 52.7 & 0.0  \\
    PE         & .00 & .00 & 51.8 & 0.0  \\
    PRM        & .00 & .00 & 33.3 & 0.0  \\
    TV-TBIP    & .00 & .00 & 45.5 & 0.0  \\
    Wordfish   & .00 & .00 & 19.1 & 0.0  \\
    Wordscores & .00 & .01 & 24.2 & 0.4  \\
    Wordshoal  & .00 & .00 & 52.2 & 0.0  \\
    \hline
  \end{tabular}
  \captionof{table}{Cleaned-profile ICC before and after per-run
    anchor-based polarity alignment, share of run groups sign-flipped by
    the alignment (among decidable groups), and share of run groups
    excluded as unalignable. For SS and CPTM, part of the
    aligned-vs-headline difference stems from restricting to alignable
    runs: on that subset, the unaligned ICCs are .36 and .06,
    respectively.}
  \label{tab:polarity}
\end{minipage}

\section{Hyperparameter Sensitivity Profiles for All Algorithms}
\label{sec:app-shap}

\begin{figure}[h]
  \centering
  \includegraphics[width=\textwidth]{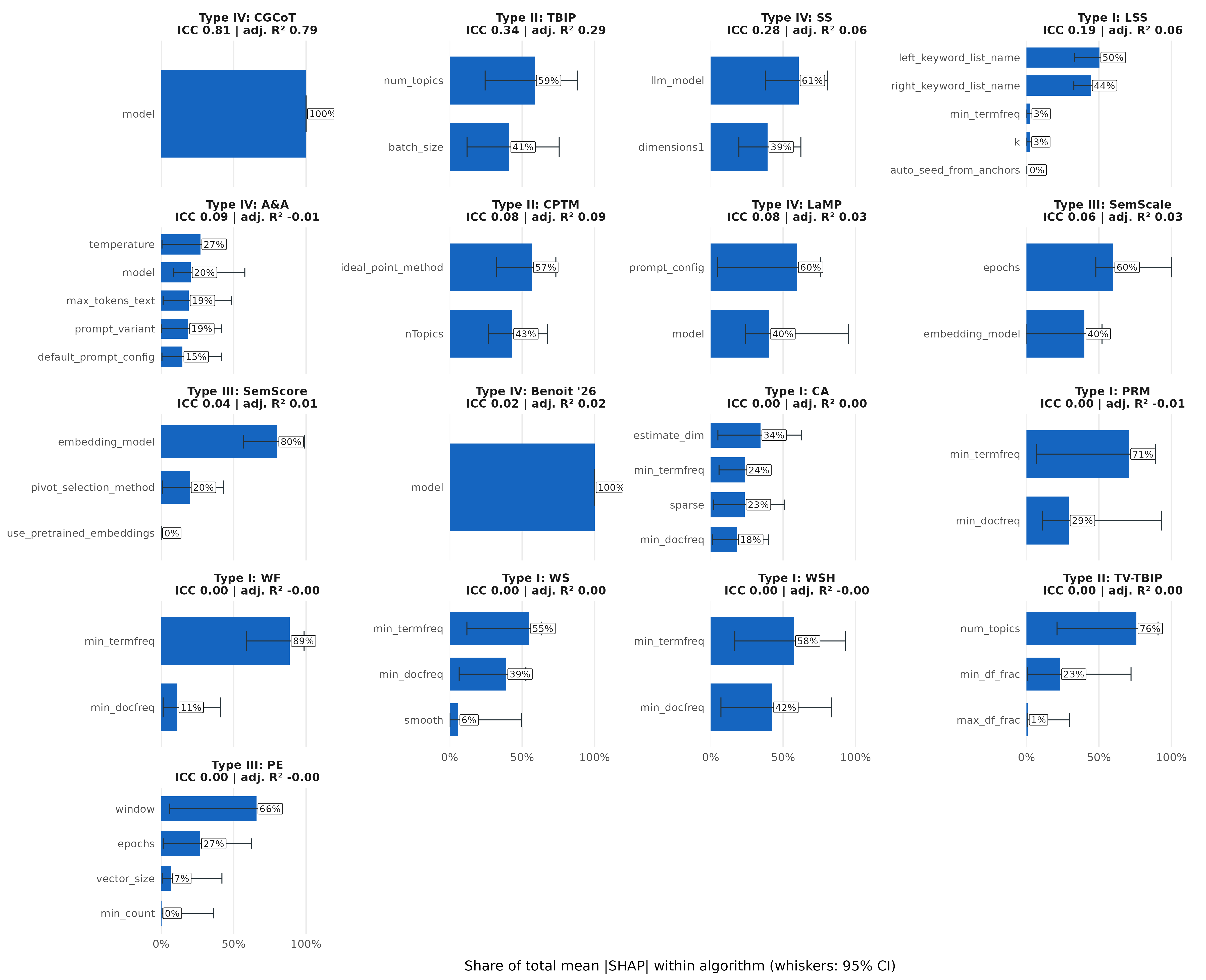}
  \caption{Within-algorithm distribution of hyperparameter sensitivity for
    all 17 algorithms (share of total mean absolute SHAP value per varied
    implementation parameter; whiskers give 95\% percentile intervals from
    the cluster bootstrap over cleaned profiles). Facets are ordered by
    the ICC in Figure~\ref{fig:icc} and additionally report the adjusted
    $R^{2}$ of that linear model. For algorithms with an ICC near
    zero or a near-zero model fit, the shares partition a negligible or
    weakly explained amount of variance and should not be substantively
    interpreted.}
  \label{fig:shap-all}
\end{figure}

\section{Party Landscapes under Alternative Researcher Choices}
\label{sec:app-landscape}

\begin{figure}[h]
  \centering
  \includegraphics[width=\textwidth]{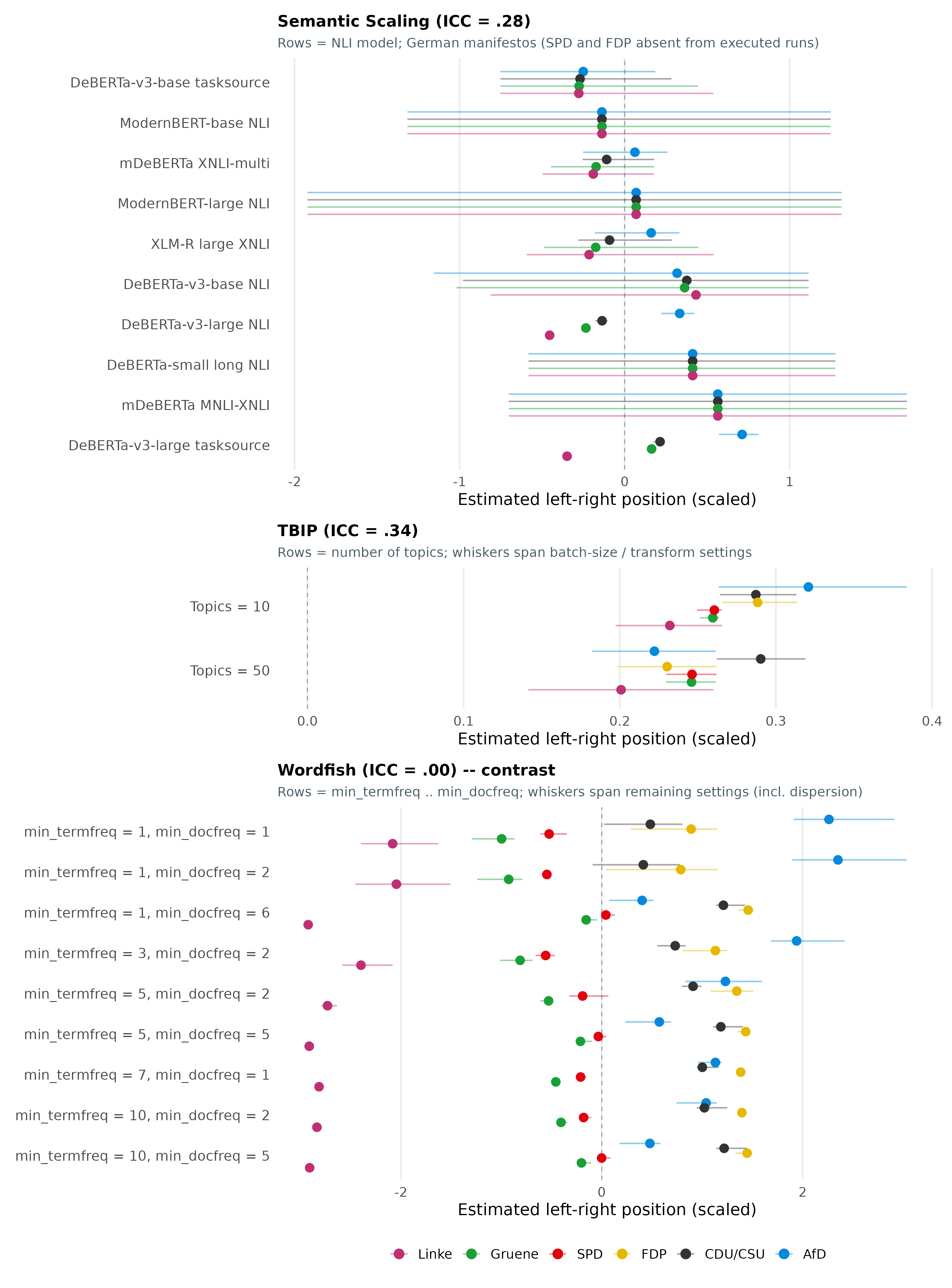}
  \caption{Party landscapes under alternative researcher choices for
    Semantic Scaling, TBIP, and Wordfish (German manifestos). Presentation
    as in Figure~\ref{fig:landscape}; row definitions differ by algorithm
    (NLI model; number of topics; \texttt{min\_termfreq} $\times$
    \texttt{min\_docfreq}). Whiskers span remaining
    within-row specification means.}
  \label{fig:landscape-app}
\end{figure}

Figure~\ref{fig:landscape-app} extends the landscape presentation of
Figure~\ref{fig:landscape} to the remaining manifesto-compatible algorithms
with non-negligible profile-level sensitivity, plus Wordfish as a contrast case
with ICC of zero. For Semantic Scaling, rows correspond to the underlying
NLI model (the dominant SHAP driver); for TBIP, to the number of topics;
for Wordfish, to the two hyperparameters with the largest SHAP shares in
Figure~\ref{fig:shap-all} (\texttt{min\_termfreq} and
\texttt{min\_docfreq}). All panels use German
manifestos. CGCoT is omitted because manifesto corpora were excluded from
its executed grid; it was evaluated on speeches and tweets only. Semantic
Scaling likewise lacks SPD and FDP in
the executed German-manifesto runs.

The three panels illustrate different forms of sensitivity. Semantic
Scaling's NLI-model choice can reorder parties and reverse the AfD's
polarity across models. TBIP's topic-count choice, despite a pooled ICC of
.34, leaves the German-manifesto party order largely intact in this cell:
high profile-level sensitivity need not imply landscape reordering in every
corpus. Wordfish's landscapes under alternative minimum term- and
document-frequency thresholds are nearly identical, consistent with its
ICC of zero.

\clearpage

\section{Online Supplementary Material}
\label{sec:app-online-supplementary}

Online supplementary material is available at
\url{https://doi.org/10.7910/DVN/LYURVY}.

The following zip files are available on the online repository:

\begin{itemize}
  \item \texttt{experiment\_pipeline\_replication\_data\_and\_code.zip}: the experimental pipeline, replication data, and code used in the experiments.
  \item \texttt{cpss\_at\_konvens\_harvard\_dataverse.zip}: the data and code for the analysis of this paper at Konvens 2026.

\end{itemize}

\section*{Acknowledgments}
The author received financial support via the Graduate Center for Doctoral
Researchers at the Bavarian Research Institute for Digital Transformation
(bidt).

\end{document}